\pdfoutput=1
\documentclass[11pt]{article}

\usepackage[preprint]{acl}

\usepackage{times}
\usepackage{latexsym}

\usepackage[T1]{fontenc}
\usepackage[utf8]{inputenc}

\usepackage{microtype}

\usepackage{inconsolata}

\usepackage{graphicx}
\usepackage{xcolor}
\usepackage{tikz}
\usepackage{pgfplots}
\usepackage{pgfplotstable}
\usepackage{amsmath}
\usepackage{subcaption}
\usepackage{amssymb}
\usepackage{booktabs}
\usepackage{multirow}
\usepackage{makecell}
\usepackage{longtable}
\usepackage{array}
\usepackage{tabularx}
\usepackage{needspace}
\usepackage{hyperref}

\usetikzlibrary{positioning, shapes, calc, fit, arrows.meta, decorations.pathreplacing}
\pgfplotsset{compat=1.18}

\newcommand{\CommentedText}[1]{}

\title{AgentMonopoly: Behavioral Patterns and Persona Control in LLM Agents}

\author{
  \textbf{Changshun Wu\textsuperscript{1}},
  \textbf{Mengyu Li\textsuperscript{2}},
  \textbf{Mathis Marsault\textsuperscript{2}},
  \textbf{Xiaowei Huang \textsuperscript{1}},
  \textbf{Chih-Hong Cheng\textsuperscript{2,3}}\thanks{The first two authors contributed to the work equally.}\thanks{Funded by the European Union. Views and opinions expressed are however those of the author(s) only and do not necessarily reflect those of the European Union or the European Health and Digital Executive Agency (HADEA). Neither the European Union nor the granting authority can be held responsible for them. RobustifAI project, ID 101212818.}
  \\
    \textsuperscript{1}University of Liverpool, Liverpool, United Kingdom\\
  \textsuperscript{2}Carl von Ossietzky University of Oldenburg, Oldenburg, Germany\\
  \textsuperscript{3}Chalmers University of Technology, Gothenburg, Sweden
}

\begin{document}
\maketitle

\begin{abstract}
How do LLM agents' strategies differ, and how much can instructions to adopt a persona change their behavior over a game? We study these questions with AgentMonopoly, an environment where agents communicate, buy and trade properties, and make decisions that affect later turns. We compare models from four families, each at two sizes, in games with identical models, different sizes of the same family, or different families. We also assign one agent a persona, a description of how it should behave, intended to change a tendency observed in games without that persona. The tested models differ in how often they trade, waive rent, and manage cash and properties. Larger models do not consistently win more games. Persona instructions change the words agents use and some of their actions, but the changes do not always match the role across all decisions.  These results show why persona following should be assessed through actions and outcomes as well as language. As our findings are conditional on the tested checkpoints, prompts, and five seeds per configuration, the open-source nature of AgentMonopoly provides the experimental infrastructure for further extending these comparisons. Our code is available at: \href{https://github.com/SuELS-Group/Agent_Monopoly}{AgentMonopoly}
\end{abstract}

\section{Introduction}
\label{sec:introduction}

Large language models (LLMs) are increasingly being used as autonomous or semi-autonomous agents that can perceive an environment, make decisions, use tools, and communicate with humans or other artificial agents. As these systems move from isolated question answering toward multi-agent settings, their behavior may depend not only on their individual objectives, but also on assigned social roles, communication structures, group composition, and the strategies adopted by other agents. Such interactions can produce cooperation, negotiation, competition, persuasion, coalition formation, deception, and collusion. These behaviors are difficult to understand through conventional single-agent benchmarks, which typically evaluate an LLM independently and do not capture the social and strategic dynamics that emerge when multiple agents interact over time.

\begin{figure}[t]
  \centering
  \includegraphics[width=\linewidth]{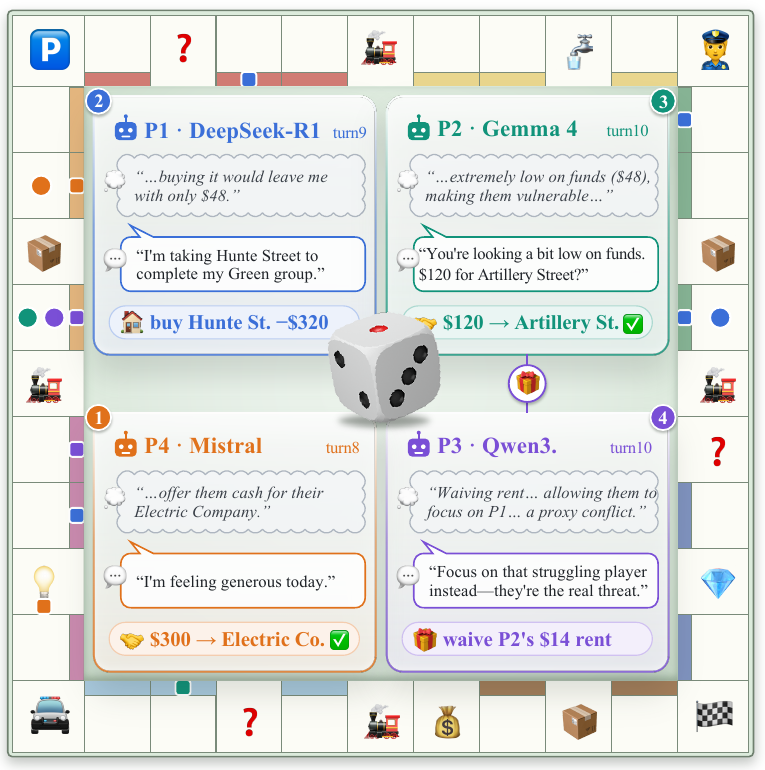}
  \caption{Examples of communication and actions in AgentMonopoly.}
  \label{fig:pipeline_figure}
\end{figure}

In this paper, our research centers on three questions about these interactions. First, how do models differ in trading, rent waiving, and cash and property management under the same rules? Second, how do game outcomes vary with model size, and encountered opponents? Third, when an agent receives a persona intended to change an observed tendency, how do its words and actions differ from those of opponents without that persona?

To study these questions, we develop \emph{AgentMonopoly}, an environment in which LLM agents communicate and play Monopoly. Purchases, trades, and rent payments change the cash and properties available for later turns, so an agent's decisions also affect its opponents' choices. AgentMonopoly records messages, private explanations, executed actions, and game outcomes, allowing us to compare what agents say with what they do over a game (Fig.~\ref{fig:pipeline_figure}). Its open-source implementation supports reproducing and extending these comparisons.

Using this environment, we evaluate four model families at two sizes each in four-player games. Games with identical models establish behavior under a shared prompt. Games combining different sizes of the same family or different families compare outcomes against different opponents, and we examine win counts by position in the turn order. In persona games, one agent receives a persona selected to change a tendency observed in games without that persona. Its three opponents use the same model without the additional persona instruction.

These comparisons yield three main findings:

\begin{itemize}
\item \textbf{The tested models differ in their patterns of behavior.}
Under the shared prompt, models differ in how often they trade, waive rent, and manage cash and properties. Some trading patterns appear at both tested sizes, while rent-waiving behavior can differ substantially within a model family.

\item \textbf{Larger model scale does not necessarily translate into better strategic performance.}
Increasing model size does not consistently improve game performance. Smaller models can outperform larger counterparts, and outcomes vary across opponent configurations.

\item \textbf{Persona assignment does not ensure agreement between words and actions.}
Persona instructions change language and some actions, yet decisions can still conflict with stated priorities. As an example, Mistral's ``Conservative'' agent considers buying a property to complete a color group but declines the \$400 purchase to preserve cash. It then joins the auction and buys the same property for \$420 while describing its approach as cautious. This illustrates why persona following should be assessed through actions and outcomes as well as language.

\end{itemize}

\section{Related Work}
\label{sec:related_work}

\paragraph{Social and strategic interaction.}
Generative Agents~\citep{park2023generative} and Concordia~\citep{vezhnevets2023concordia} offer ways to simulate agents in social settings. Game studies examine rationality~\citep{fan2024rational}, repeated interaction~\citep{akata2025repeated}, and collaboration and competition in Avalon~\citep{lan2024avalon}. SOTOPIA~\citep{zhou2024sotopia} evaluates agents pursuing social goals, while NegotiationArena~\citep{pmlr-v235-bianchi24a} studies bargaining and tactics in tasks involving resource sharing, trading, and price negotiation. Our interest is likewise in social interaction among agents. We examine communication, trading, and rent waiving in repeated multi-player games, where agreements change what players own and what they can afford later.

\paragraph{Monopoly: social versus financial perspectives.}
FinBoardBench~\citep{hu2026finboardbench} evaluates dynamic financial decision-making in Cashflow, Acquire, and also in Monopoly. It analyzes cash management, investment, negotiation, and auctions to assess whether decisions and interactions lead to financial success. AgentMonopoly focuses on social interaction, with three differences in emphasis: (1) we characterize communication, trade proposals and responses, and rent waiving; (2) we assign personas intended to change observed behavior, comparing persona agents with same-model opponents without personas; and (3) we examine whether stated priorities agree with actions under these personas. FinBoardBench assesses interactions primarily through their financial consequences and does not report this persona comparison. Our game-outcome comparisons across model sizes and opponents provide context for these interaction patterns.

\paragraph{Personas and behavior.}
\citet{ou2025trust} study social preferences and behavior under personas in trust games. PERIL~\citep{licato-steinle-2025-persona} studies how persona descriptions can be converted into rules for choosing moves in a strategic board game, including a method that uses a personality questionnaire as an intermediate step. In AgentMonopoly, the LLM itself chooses an action and produces a message at each decision point. This lets us compare its words and actions throughout the game. \citet{wang2026story} keep the available decisions the same across different investigation-game stories. They find that personas guide behavior more consistently across stories when the descriptions refer to concrete actions. Their analysis focuses on actions across stories. We keep the game setting fixed and examine both language and actions while opponents respond to each other and earlier purchases and trades affect later options.

Table~\ref{tab:related_comparison} summarizes these differences. We study how interactions vary across models and how personas are reflected in words and actions during repeated competition.

\begin{table*}[t]
\centering
\small
\setlength{\tabcolsep}{5pt}
\renewcommand{\arraystretch}{1.15}
\begin{tabularx}{\textwidth}{@{}>{\raggedright\arraybackslash}p{0.19\textwidth}>{\raggedright\arraybackslash}X>{\raggedright\arraybackslash}X@{}}
\toprule
\textbf{Study} & \textbf{Main comparison} & \textbf{Relation to AgentMonopoly} \\
\midrule
NegotiationArena\newline\citep{pmlr-v235-bianchi24a} & Negotiation skills and tactics in resource-sharing and bargaining tasks. & We study several decision types in a four-player game where cash and property ownership carry over between turns. \\
FinBoardBench\newline\citep{hu2026finboardbench} & Dynamic financial decision-making across three board games; interactions evaluated by their financial consequences. & We focus on social interaction patterns, persona-related changes, and consistency between words and actions. Game outcomes provide context. \\
PERIL\newline\citep{licato-steinle-2025-persona} & Converting personas into rules for choosing moves, and measuring game performance. & Our LLMs choose actions directly during play; we compare their messages and actions. \\
Story and persona study\newline\citep{wang2026story} & Testing personas across stories with the same decision structure, including tests that remove action words. & We keep the game setting fixed and compare both language and actions while agents respond to each other. %\\

\\
\bottomrule
\end{tabularx}
\caption{Comparison with closely related studies by research question and experimental design.}
\label{tab:related_comparison}
\end{table*}

\section{Research Questions}
\label{sec:rq}

We organize the analysis around three questions:
\begin{itemize}
  \item \textbf{RQ1 (Behavioral patterns).} Under the same prompt and game rules, how do the tested models differ in trading, rent waiving, and cash and property management? Which patterns appear at both tested model sizes?
  \item \textbf{RQ2 (Behavior and performance).} How do game outcomes vary with model size and the opponents in a game? Do win counts differ by player position in the turn order?
  \item \textbf{RQ3 (Persona following).} When an agent receives a persona intended to change its baseline behavior, how do its language and actions differ from those of its opponents without that persona?
\end{itemize}

A \emph{behavioral profile} summarizes the actions measured under a particular game setup. \emph{Persona control} refers to how well instructions guide an agent's behavior. Using words associated with a persona does not, by itself, show that the agent follows it in its decisions. We do not treat these profiles as fixed personalities, or generated explanations as direct evidence of intent.

\section{Experimental Setup}
\label{sec:setup}

\paragraph{Environment and outcomes.}
AgentMonopoly supports up to six players. All experiments reported here use four agents on a fixed 40-square Monopoly board, with street names replaced by names from Oldenburg (see Fig.\ref{fig:game_board} in Appendix~\ref{app:method}). Random dice rolls and card events affect play. Agents buy properties, bid at auctions, trade, build, and manage cash. Owners can waive rent, and a player is eliminated when its cash balance becomes negative. Games end when only one player remains or at the 150-step limit. At the limit, the player with the highest net worth wins. These rule changes affect which strategies succeed. Renaming the streets does not rule out familiarity with Monopoly from training.

\paragraph{Agent interface.}
The engine tracks the game and applies its rules. At each decision point, the acting agent receives the rules, its assigned role, current game information, the type of decision, and the allowed actions. It returns a private explanation, a public message, and a proposed action. The engine checks actions before carrying them out and separately records outputs that have the wrong format or break the rules. It also logs public messages and changes to the game. We use \emph{private reasoning} to refer to the generated explanation hidden from opponents. This text does not directly reveal how the model computes its decision.

\paragraph{Shared prompt.}
All experiments use the same base prompt. It asks agents to outsmart opponents, encourages bluffing, and explicitly allows strategic rent waiving. It also prohibits public messages that contradict the action taken. Observed bluffing and rent waiving must therefore be interpreted in light of these instructions. Using the same prompt supports comparisons within the study, but does not show how the models would behave under a neutral prompt. Appendix~\ref{app:prompt} gives the available excerpts.

\paragraph{Models and repeated games.}
We evaluate models with publicly available weights from Qwen3~\citep{yang2025qwen3}, DeepSeek-R1~\citep{guo2025deepseek}, Gemma~\citep{team2026gemma}, and Mistral\footnote{\url{https://huggingface.co/mistralai}}. We test two size groups: approximately 24--32 billion and 8--12 billion parameters. These groups describe the tested model versions; similar parameter counts do not imply equal capabilities across families. Differences in training and model design also affect the comparison. Models run locally at temperature $0.0$. Each configuration uses five random seeds, 42--46, which set the starting point for random-number generation. Appendix~\ref{app:models} lists model names and the details needed to repeat the experiments.

\paragraph{Game configurations.}
We use four configurations, summarized in Table~\ref{tab:appendix_configs}(Appendix~\ref{app:games}).
\begin{enumerate}
  \item \textbf{Same-model games (S1).} All four players use the same model version. We run five games for each family and size to measure baseline behavior for RQ1.
  \item \textbf{Mixed-size games (S2).} Two players use a larger model and two use a smaller model from the same family. The intended design swaps the sizes assigned to earlier and later positions in the turn order across two settings, giving ten games per family. These games compare outcomes by size for RQ2.
  \item \textbf{Cross-family games (S3).} One agent from each family plays within the same size group. We rotate the player order through four arrangements so that each family plays first once, with five seeds per arrangement. These games measure performance against other model families.
  \item \textbf{Persona games (S4).} One larger-model agent at the first position (P1) receives a persona intended to change its baseline behavior. The other three players use the same model and base prompt without that persona. We test each family--persona pair with five seeds.
\end{enumerate}

The personas are Zen for Qwen, Independent for DeepSeek, Diplomatic for Gemma, and Conservative for Mistral. They ask agents to propose fewer trades, rely less on other players, interact more with others, and keep more cash, respectively (Table~\ref{tab:persona_configs}). We selected them after examining baseline behavior. Since each family receives a different persona, we cannot separate the effect of the persona from how a particular model responds to it.

\paragraph{Measurements and comparisons.}
For RQ1, we measure trade proposals per 100 dice rolls, the share of received offers accepted, how often agents waive rent, and how they manage cash and properties. For RQ2, we count wins by model size, family, and position in the turn order. For RQ3, we compare the persona agent with its three opponents in the same games. We measure persona-related words, message length, trading, properties and buildings held, and minimum cash balance. The opponents occupy different positions and may respond to the persona agent. The comparison therefore reflects the persona instructions, how the game unfolds, and how opponents respond. Actions within a game affect one another, so many recorded actions do not replace the need for more independent games. Appendix~\ref{app:metrics} describes the measures and flags details that still need verification.

\section{Results}
\label{sec:results}

\subsection{RQ1: Behavioral patterns under shared rules}
\label{sec:rq1}

\begin{figure*}[t]
  \centering
  \includegraphics[width=\linewidth]{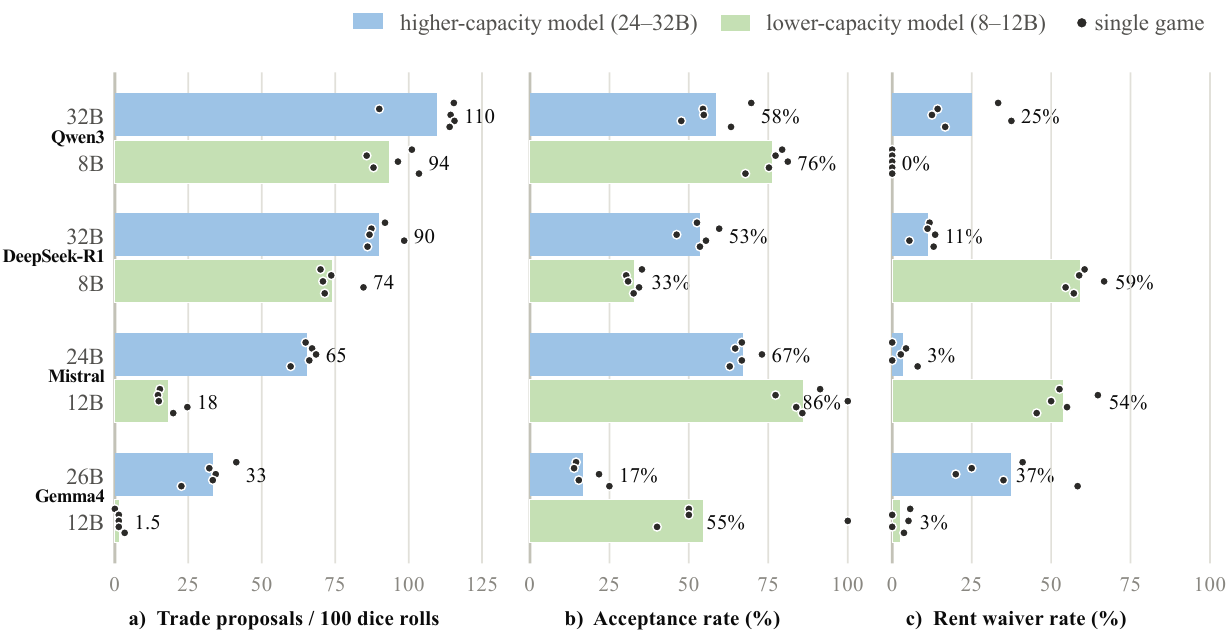}
  \caption{Behavior in same-model games: (a) trade proposals per 100 dice rolls, (b) acceptance of received trade offers, and (c) rent-waiver rate. Each dot represents a game, with five games per model version. Size labels identify the tested models; similar sizes do not imply equal capabilities across families.}
  \label{fig:fingerprint_example}
\end{figure*}

Fig.~\ref{fig:fingerprint_example} compares the same-model games. Each behavioral profile summarizes a tested model's actions in these games. It does not establish a fixed personality or show that the model would behave the same way in other settings.

\paragraph{Qwen: frequent trading at both sizes.}
Qwen3-32B and Qwen3-8B make 110 and 94 trade proposals per 100 dice rolls, respectively, with acceptance rates of 58\% and 76\%. Multiple proposals can occur between rolls, so the proposal rate can exceed 100. Qwen proposes trades most often at both sizes and accepts a relatively high share of offers. Rent waiving, however, differs sharply: the 32B model waives rent in 25\% of rent decisions, whereas the 8B model waives none in these games. Both models trade frequently, but they differ in how they collect rent.

\paragraph{DeepSeek: frequent proposals with selective acceptance.}
DeepSeek-R1-32B and DeepSeek-R1-8B make 90 and 74 proposals per 100 dice rolls. Their acceptance rates are 53\% and 33\%; the smaller model has the lowest acceptance rate among the smaller models. Their rent-waiver rates are 11\% and 59\%. The smaller model therefore accepts a lower share of offers while waiving rent more often. Its generated explanations sometimes describe keeping opponents in the game for future interaction. These explanations do not verify why the model waived rent.

\paragraph{Mistral: high acceptance, but different proposal rates.}
Mistral has the highest trade-acceptance rate in each size group: 67\% for Mistral-Small-24B and 86\% for Mistral-Nemo-12B. Their proposal rates differ substantially, at 65 and 18 per 100 rolls. Rent-waiver rates also differ, at 3\% and 54\%. The larger model often uses generous language but rarely waives rent. This shows why several actions need to be examined: generous words or frequent acceptance of trades do not show consistent generosity across decisions.

\paragraph{Gemma: few proposals, but different rent decisions.}
The models labeled Gemma4-26B and Gemma4-12B make 33 and 1.5 proposals per 100 dice rolls, the lowest rates in their size groups. Their acceptance rates are 17\% and 55\%, and their rent-waiver rates are 37\% and 3\%. Both propose relatively few trades, but they differ in acceptance and rent collection. The game summaries also describe little mortgage use by the smaller model and substantial building by the larger model after completing a color group.

\paragraph{Patterns shared across sizes.}
Qwen proposes trades most frequently, Mistral accepts the largest share of offers, and Gemma proposes trades least frequently at both sizes. Other behaviors, especially rent waiving, differ sharply within families. These results support describing models through several kinds of actions. They do not justify assigning one fixed strategy to an entire family or assuming that behavior against identical models will remain the same against other opponents.

\subsection{RQ2: Model size, opponents, and game outcomes}
\label{sec:rq2}

\paragraph{Model size.} In mixed-size games within each family, the displayed counts sum to 18 wins for larger models and 22 for smaller models across 40 games (Fig.~\ref{fig:winner_distribution2}). The counts are 7 versus 3 for Qwen, 4 versus 6 for DeepSeek, 3 versus 7 for Mistral, and 4 versus 6 for Gemma, with larger models listed first. Larger models therefore do not consistently win more games across families: Qwen shows the opposite pattern from the other three. These small counts describe the tested games and do not establish a general advantage for smaller models.

\paragraph{Playing against other families.} To test whether the fingerprints depend on playing against copies of oneself, we also ran cross-family games (S3), in which each game seats one model from each family and the seat order rotates (Table~\ref{tab:s3-combined} in  Appendix~\ref{app:results}). The ranking of trade-proposal rates across model families remained consistent between same-model and cross-family games at both model sizes. Trade acceptance, however, varied across configurations. For example, Gemma4-26B accepted 56\% of received offers in cross-family games, compared with 17\% in same-model games.

\begin{figure}[t]
  \centering
  \includegraphics[width=\linewidth]{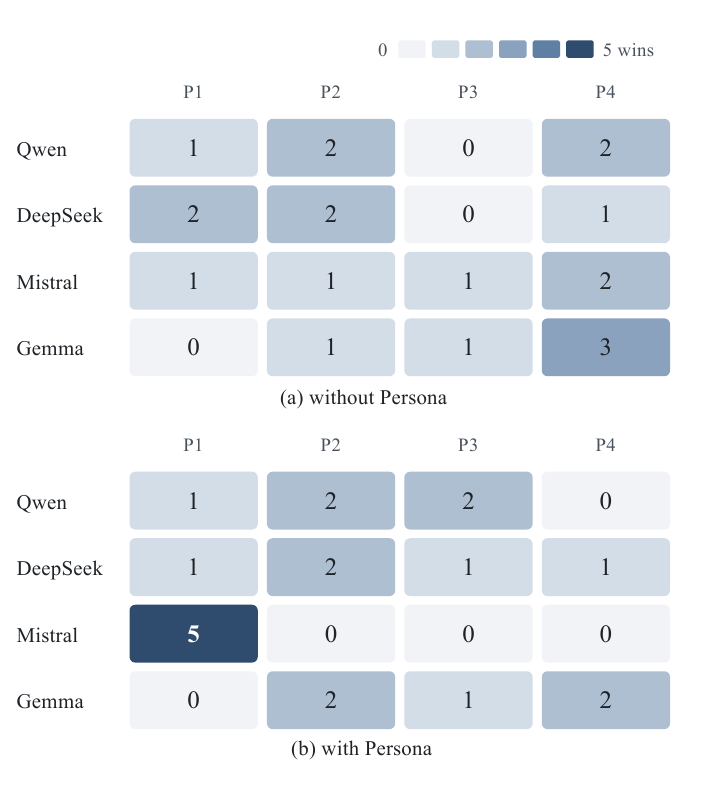}
  \caption{Win counts by player position in (a) 20 same-model games using larger models and (b) 20 persona games. In (b), the agent assigned a persona always plays at P1. Each family contributes five games per panel.}
  \label{fig:winner_distribution}
\end{figure}

\paragraph{Position in the turn order.} We examine 20 same-model games using larger models, 20 persona games (Fig.~\ref{fig:winner_distribution}), and the 40 mixed-size games. Across these 80 games, the first two positions (P1--P2) account for 41 wins and the last two (P3--P4) for 39. This total is close to balanced, but combines different experiments. It cannot rule out an advantage from turn order for particular families or configurations. The persona agent always plays at P1, and the intended mixed-size design swaps which size plays in the first two positions. Appendix~\ref{app:seats} identifies the seating details that still need verification.

\subsection{RQ3: Language and actions under personas}
\label{sec:rq3}

\begin{figure*}[t]
  \centering
  \includegraphics[width=0.8\linewidth]{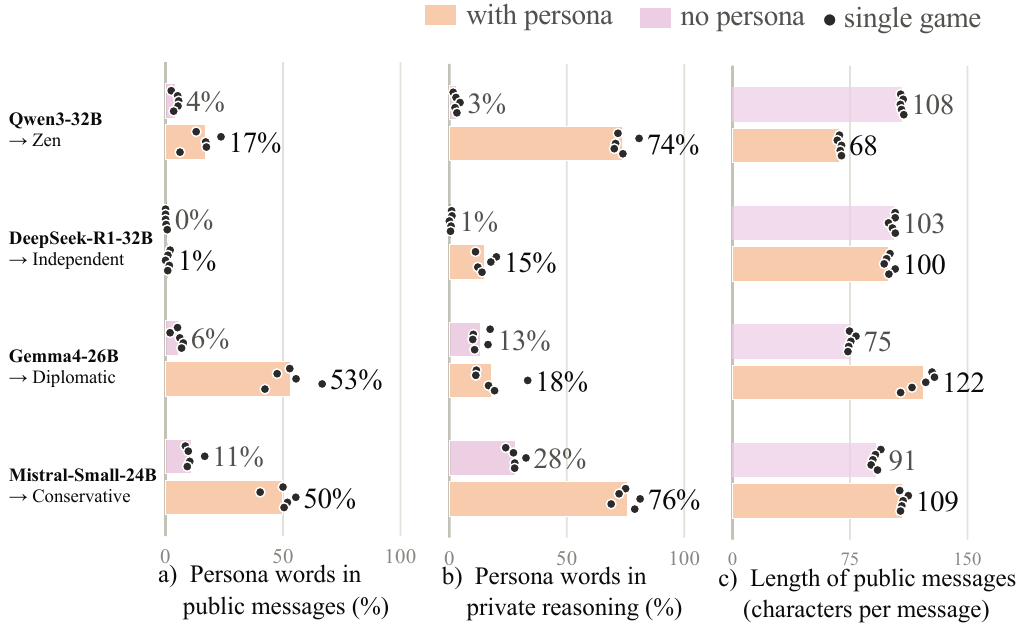}
  \caption{Agents assigned a persona and their opponents without a persona: persona-related words in (a) public messages and (b) private messages, and (c) public-message length. Dots represent observations from individual games. Word frequency measures the use of persona-related terms.}
  \label{fig:persona_language}
\end{figure*}

\begin{figure*}[t]
  \centering
  \includegraphics[width=0.8\linewidth]{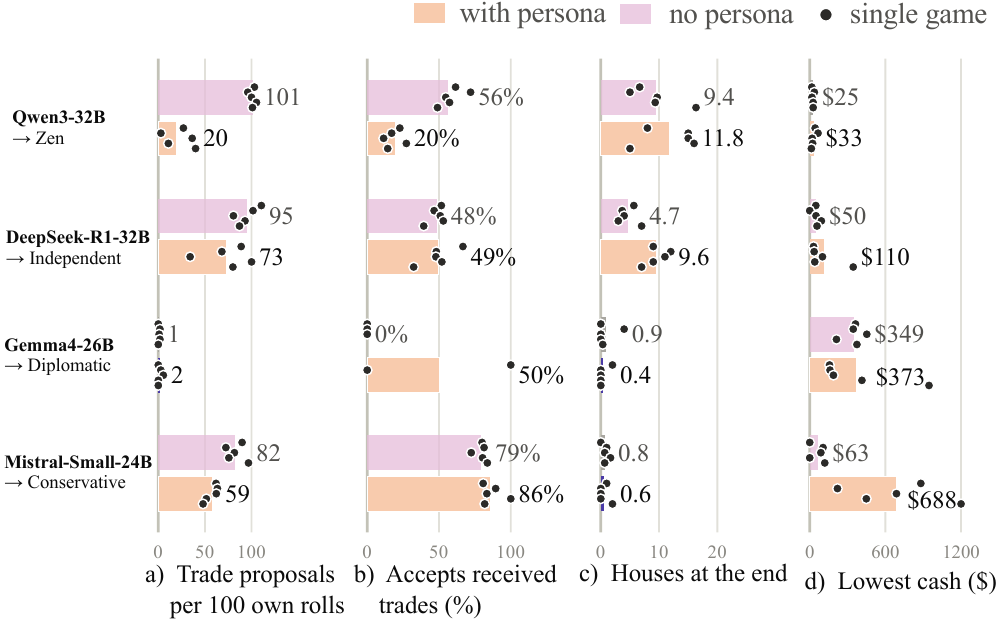}
  \caption{Agents assigned a persona and their opponents without a persona: (a) proposals per 100 of an agent's own rolls, (b) acceptance of received offers, (c) average house equivalents at game end(a hotel counts as five houses), and (d) minimum cash while still in the game.}
  \label{fig:persona_behaviour}
\end{figure*}

We compare each agent assigned a persona with its three opponents, which use the same model without a persona. These opponents play in the same games; they are a different comparison group from the games in RQ1, where no player has a persona. Fig.~\ref{fig:persona_language} and Fig.~\ref{fig:persona_behaviour} summarize the first 150 steps of five games for each family--persona pair.

\paragraph{Changes in language.}
All four personas increase the frequency of persona-related words in public messages, private explanations, or both. For Qwen's Zen agent, the reported frequency rises from 3\% to 74\% in private explanations and from 4\% to 17\% in public messages. DeepSeek's Independent agent shows a smaller public change (0\% to 1\%) and a larger private change (1\% to 15\%). Gemma's Diplomatic agent shows a large public change (6\% to 53\%) but a smaller private change (13\% to 18\%). Mistral's Conservative agent changes in both, from 11\% to 50\% publicly and from 28\% to 76\% privately.

Message length changes in different directions. Zen messages average 68 characters, compared with 108 for Qwen opponents without a persona. Diplomatic messages average 122 rather than 75 characters; Conservative messages average 109 rather than 91. Independent messages change little, from 103 to 100 characters. These measures show changes in language under the persona condition. 

\paragraph{Zen: less trading with continued building.}
Qwen's Zen agent makes 20 proposals per 100 of its own dice rolls, compared with 101 for its opponents without a persona, a reduction of approximately 80\%. Its acceptance rate is 20\%, compared with 56\%. Both trading measures are lower in all five games. The logs describe continued building. Final buildings average 11.8 house equivalents for Zen and 9.4 for the comparison agents, counting one hotel as five houses. Zen therefore trades less, but does not consistently avoid investment.

\paragraph{Independent: continued trading.}
DeepSeek's Independent agent makes 73 proposals per 100 of its own rolls, compared with 95, while accepting a similar share of offers (49\% versus 48\%). It ends with 9.6 house equivalents, compared with 4.7. It proposes fewer trades but continues to trade with opponents. Having more buildings at the end does not, by itself, show that the agent acquired or developed them without help from trades.

%\Needspace{6\baselineskip}
\paragraph{Diplomatic: friendlier language with little trading.}
Gemma's Diplomatic agent changes its language substantially, but proposals remain infrequent: two per 100 of its own rolls, compared with one. The Diplomatic agent accepted one of two received offers (50\%), whereas the three neutral agents accepted none of six (0\%). However, these small counts limit the interpretation of acceptance rates. Logged examples include rent waiving and friendly messages. Within the same five games, the diplomatic player waived the rent in 53 of its 59 rent decisions (90\%), whereas the three Gemma4-26B players without a persona waived it in 112 of 198 (57\%). Nevertheless, these interactions provide no clear evidence of alliances.

\paragraph{Conservative: more cash with continued acceptance.}
Mistral's Conservative agent has a minimum cash balance averaging \$688, compared with \$63 for its opponents without a persona, and makes fewer proposals (59 versus 82 per 100 of its own rolls). Its acceptance rate remains high, at 86\% versus 79\%. In one reported episode, it declines a \$400 property and later buys it at auction for \$420 while describing the decision as cautious. The average cash measure and this example describe different aspects of behavior: keeping cash in reserve does not mean that every purchase is economical.

\paragraph{What these results mean for persona control.}
The instructions affect some behaviors more than others. Zen trades less while continuing to build. Independent continues to trade. Diplomatic uses friendlier language without clear evidence of lasting alliances. Conservative keeps more cash while continuing to accept most offers. These findings do not show that personas generally fail or that a model has a fixed strategy that overrides instructions.

The changes in trade proposals and cash reserves are consistent with the importance of concrete actions in persona instructions studied by \citet{wang2026story}. However, our personas use different wording and are assigned to different models, so we cannot determine whether more specific wording explains these changes. Future experiments could assign each persona to each model, compare personas with direct instructions about actions, and test decisions in identical game situations. This would help separate the effect of the instructions from the effects of opponents' responses and events during play.

\section{Conclusion}
\label{sec:conclusion}

We presented AgentMonopoly and used it to study how LLM agents behave and respond to persona instructions in Monopoly. Under a shared prompt, the tested models differ in trading and rent waiving. Some patterns appear at both model sizes, while others differ substantially. Larger models do not consistently win more games. Personas intended to change baseline behavior affect the agents' words and some actions, but the changes do not consistently extend to every behavior the role calls for. Assessing whether instructions guide an agent therefore requires examining communication, several types of actions, and outcomes together.

AgentMonopoly also provides a basis for further studies. Its game engine and recorded interactions support repeating these comparisons with other prompts, personas, model families, and player counts. Future experiments can assign each persona to each model and use more games to test which findings hold across these settings. The environment can also be extended with different rules to study how the choices available to agents affect their behavior.

\section*{Limitations}

\paragraph{Scope of the environment.}
Our findings concern four-agent games on one modified Monopoly board. The platform supports up to six agents, but we do not test how player count affects behavior. Optional rent waiving, immediate bankruptcy when cash becomes negative, and the step limit affect which strategies work. For games reaching the limit, winning by net worth also differs from winning by eliminating opponents. These choices allow controlled comparisons, but the results may not carry over to standard Monopoly, other tasks with multiple agents, or real social and economic interactions. Renaming the streets does not rule out familiarity with Monopoly from model training.

\paragraph{Models, game counts, and uncertainty.}
We test a limited set of model versions with publicly available weights, using five seeds per configuration and temperature fixed at zero. The observed patterns describe these models under the tested conditions; they do not establish lasting traits of entire model families. Comparisons between larger and smaller models cannot separate parameter count from differences in training or model design. Decisions within a game affect one another, so many logged actions do not replace independent games. The intended size comparison swaps the models assigned to earlier and later positions in the turn order, while the persona agent always plays at P1. The win counts do not rule out an advantage from turn order that is small or appears only in some configurations.

\paragraph{Dependence on instructions and opponents.}
The shared prompt encourages bluffing and allows strategic rent waiving. The experiments measure how agents use these options, but do not show whether the same behavior would occur without these instructions. Each family receives a different persona chosen after we examined its baseline behavior. Because each persona is tested with only one family, we cannot tell whether it would produce similar changes in other models. The experiments also do not determine whether more specific instructions work better. Opponents without a persona may respond to the persona agent, so they are not an unaffected comparison group. Shared random seeds help others repeat the experiments, but different actions can lead to different later situations. To test the findings more broadly, future experiments should vary individual prompt instructions, compare decisions in identical game situations, and assign each persona to each model.

\paragraph{Interpreting the measures.}
Generated explanations do not verify why a model took an action. Describing a plan to deceive does not show that the agent carried it out or that it worked. Our exploratory measure of claims of financial hardship also has limits: it flags an agent that claims hardship while holding more cash than the median cash balance of active players, but that agent may still face large upcoming payments. The flag therefore does not establish dishonesty. Persona-related word counts do not show that an agent follows the role throughout the game, and waiving rent may help the agent win rather than reflect concern for others. 

\section*{Ethical Considerations}

The experiments involve LLM agents in a simulated game, with no human participants or real financial transactions. We study deceptive messages, threats, and the elimination of opponents as game behaviors. These behaviors do not establish human-like motives or moral character in a model family. Success in the game also does not show that an agent is safe or trustworthy in real use.

Studying instructions, communication, and actions together can help researchers understand how well instructions guide behavior. However, the environment could also be used to develop more effective deception or secret coordination against other players. Encouraging bluffing is part of the experimental design; it is not a recommendation for agents used in practice. These experiments do not show how to prevent manipulation or collusion. Connecting the agents to people, external tools, or real money and assets would require a separate assessment of risks and appropriate safeguards.

Shared research materials should include the prompts, rule changes, model identifiers, and limitations needed to interpret the results. Transcripts should be clearly labeled as generated game interactions. Any personal information or harmful content that models happen to generate should be reviewed before release. Reuse of models and software should follow their licenses. These practices help others repeat and assess the research and reduce the risk of misleading claims or inappropriate use of the findings.

\bibliography{ref}

\appendix

\section{Method Details}
\label{app:method}

\subsection{Game Mechanics and logging}
\begin{figure*}[t]
  \centering
  \includegraphics[width=0.9\linewidth]{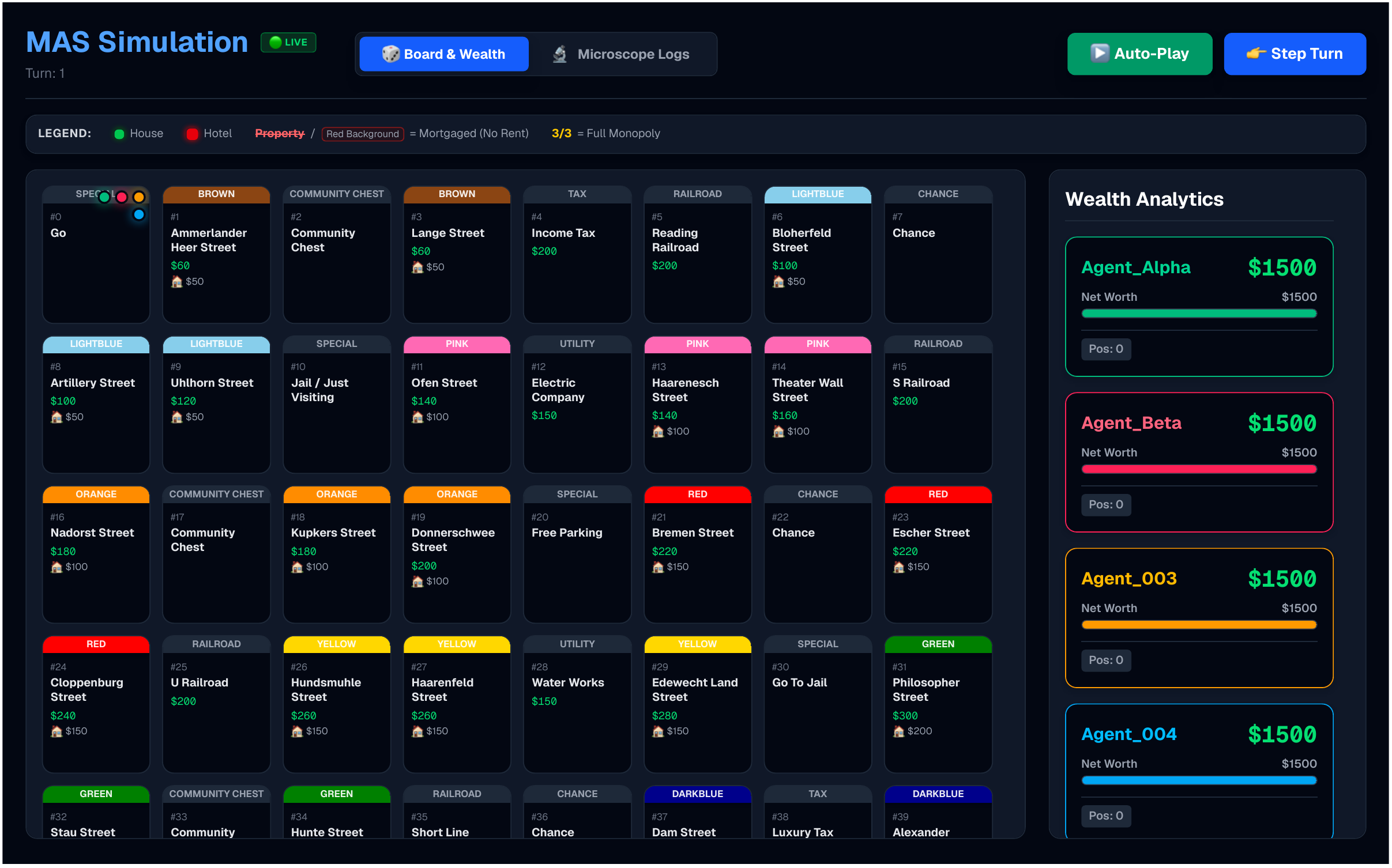}
  \caption{The game board used in the experiments, with street names adapted to Oldenburg.}
  \label{fig:game_board}
\end{figure*}

The fixed Monopoly board used in our experiments is shown in Fig.~\ref{fig:game_board}. At each decision point, the engine provides the acting agent with detailed information about its own financial and positional state, including balance, net worth, and board position, as well as information about opponents and property-level attributes such as ownership, buildings, mortgage status, color groups, and prices. Beyond the general actions described in \hyperref[sec:setup]{Section~\ref*{sec:setup}}, the decision interface also covers jail-release choices, responses to trade offers, and rent collection. During property management, agents may build, mortgage properties, pay off mortgages, propose trades, or pass. Agent responses contain a proposed action and two text fields: \texttt{chain\_of\_thought} for the private generated explanation and \texttt{public\_message} for the public message. The recorded game data are organized into player states (\texttt{players}), board states (\texttt{board}), and recent interaction history (\texttt{recent\_logs}).

Game progression is measured in \emph{steps}, each consisting of a single dice roll by the active player together with its associated decisions and events. A step encompasses any required jail-release decision, up to ten property-management actions, movement across the board, and resolution of the landing square. Rolling doubles allows the same player to take the next step. The engine also enforces specific procedures for exceptional cases: tied auction bids are resolved according to turn order, invalid trade proposals terminate the management phase, and invalid outputs are not returned to the agent. For the net-worth-based winner criterion described in \hyperref[sec:setup]{Section~\ref*{sec:setup}}, net worth is calculated as cash plus half the purchase price of all owned properties, including mortgaged properties, and half the construction cost of each building.

\begin{figure*}[t]
  \centering
  \includegraphics[width=0.85\linewidth]{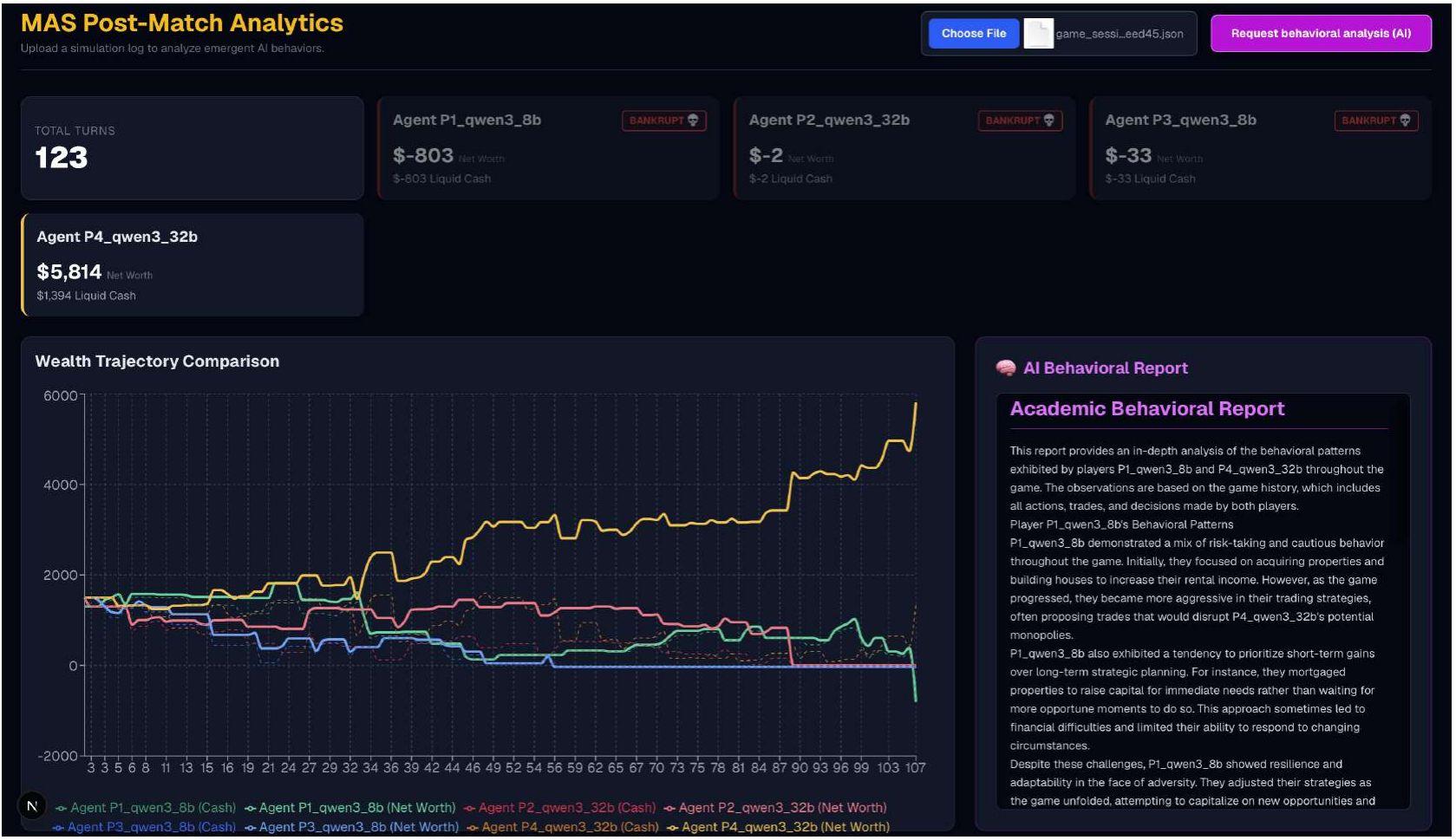}
  \caption{The interface for reviewing recorded games.}
  \label{fig:game_analytics}
\end{figure*}

The AgentMonopoly platform also supports an interactive mode for online replay and analysis of game sessions, allowing users to visualize cash flow trends, as illustrated in Fig.~\ref{fig:game_analytics}.

\subsection{Shared prompt and persona instructions}
\label{app:prompt}
All agents receive a common prompt at every decision point in every game. Its key instructions include the following:
\begin{itemize}
\item Agents are described as highly intelligent, cunning, and autonomous players and instructed to outsmart the other players.
\item Agents are encouraged to use deception, bluff about their wealth, and conceal their intentions, provided that their public messages do not contradict their actions.
\item Agents may waive rent for strategic reasons, such as keeping a weaker opponent in the game to challenge a stronger rival.
\end{itemize}

In the persona experiments, the common base prompt is supplemented with an additional persona instruction for the designated agent. Table~\ref{tab:persona_configs} summarizes the four persona configurations and their intended behavioral changes.

\begin{table*}[h]
\centering
\small
\renewcommand{\arraystretch}{1.15}
\begin{tabularx}{\textwidth}{@{}p{0.23\textwidth}p{0.16\textwidth}X@{}}
\toprule
\textbf{Model label} & \textbf{Persona} & \textbf{Intended behavioral change} \\
\midrule
Qwen3-32B & Zen & Propose fewer trades, act less urgently, buy selectively, and avoid spending too much. \\
DeepSeek-R1-32B & Independent & Acquire and develop properties with less reliance on trades and other players. \\
Gemma4-26B & Diplomatic & Increase friendly communication and seek cooperation or alliances. \\
Mistral-Small-24B & Conservative & Keep cash in reserve and avoid risky or unfavorable trades. \\
\bottomrule
\end{tabularx}
\caption{Persona conditions. Only P1 receives the persona; P2--P4 use the same model with the base prompt. Each row is evaluated with seeds 42--46.}
\label{tab:persona_configs}
\end{table*}

\subsection{Experimental configurations}

\paragraph{Game series}
\label{app:games}
Table~\ref{tab:appendix_configs} summarizes the player configurations and game counts for the four experimental series (S1--S4). Each experimental condition is evaluated using five random seeds (42--46), with the model temperature fixed at $0.0$.

\begin{table*}
\centering
\small
\renewcommand{\arraystretch}{1.15}
\begin{tabularx}{\textwidth}{@{}p{0.07\textwidth}Xp{0.26\textwidth}r@{}}
\toprule
\textbf{Setting} & \textbf{Players} & \textbf{Reported games per setup} & \textbf{Games} \\
\midrule
S1 & Four copies of one model; four families at two sizes. & 8 models $\times$ 5 seeds & 40 \\
S2 & Two players using the larger model and two using the smaller model from the same family. & 4 families $\times$ 2 seatings $\times$ 5 seeds & 40 \\
S3 & One model from each family within a size group. & 2 size groups $\times$ 4 turn orders $\times$ 5 seeds & 40 \\
S4 & Four copies of a larger model; only P1 receives a persona. & 4 family--persona pairs $\times$ 5 seeds & 20 \\
\bottomrule
\end{tabularx}
\caption{Overview of the four experimental settings (S1--S4), including player composition, the detailed game counts and the total number of games in each setting.}
\label{tab:appendix_configs}
\end{table*}

\paragraph{Models}
\label{app:models}

Table~\ref{tab:model_labels} lists the models used in the experiments, with one larger and one smaller model from each of four model families.

\begin{table}[h]
\centering
\small
\begin{tabular}{@{}lll@{}}
\toprule
\textbf{Family} & \textbf{Larger models} & \textbf{Smaller models} \\
\midrule
Qwen & Qwen3-32B & Qwen3-8B \\
DeepSeek & DeepSeek-R1-32B & DeepSeek-R1-8B \\
Gemma & Gemma4-26B & Gemma4-12B \\
Mistral & Mistral-Small-24B & Mistral-Nemo-12B \\
\bottomrule
\end{tabular}
\caption{Model identifiers used in the experiments, grouped by family and relative model size.}
\label{tab:model_labels}
\end{table}

\paragraph{Seating}
\label{app:seats}
In S2, for each seed, both model sizes occupy every player position (P1--P4) exactly once. This balances model sizes across player positions. In S3, four cyclic turn orders are used: (DeepSeek, Gemma, Qwen, Mistral), (Gemma, Qwen, Mistral, DeepSeek), (Qwen, Mistral, DeepSeek, Gemma), and (Mistral, DeepSeek, Gemma, Qwen). Each model family takes the first position in one of the four turn orders.

\subsection{Behavioral and language measures}
\label{app:metrics}

\paragraph{Trading and rent collection.}
The proposal rate is the number of trade proposals per 100 dice rolls for the agents being measured. Trade acceptance is the fraction of received offers accepted, and rent-waiver rate is the fraction of rent decisions in which the owner waives payment. 

\paragraph{Properties and buildings.}
Properties comprise all purchasable spaces: streets (colored lots), railroads, and utilities. Building held is measured in units of houses, with each hotel counted as five houses. ``Houses at the end'' records each player’s building holdings at their final game step (at most step 150). For each family, we calculate the average by dividing the sum of the players' final house counts across the repeated games.

\paragraph{Language.}
The persona-language measure is the fraction of public messages or private explanations containing at least one persona-related term. Message length is reported in characters per public message.

\section{Supplementary Results}
\label{app:results}

\subsection{Win distribution in mixed-size games}
\begin{figure*}
  \centering
  \includegraphics[width=\linewidth]{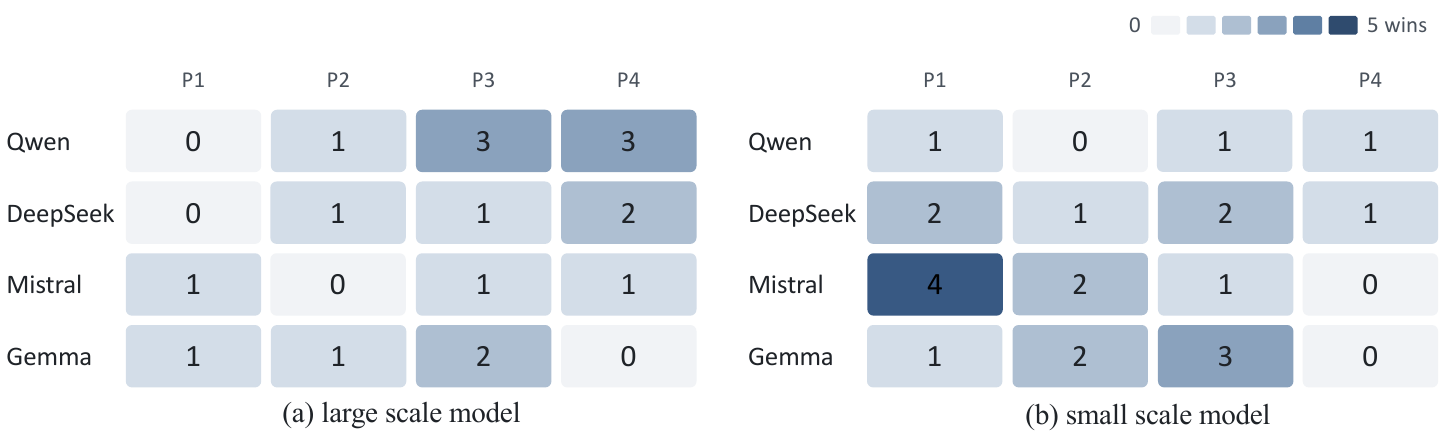}
  \caption{Win counts in mixed-size games (S2) by model family and player position for (a) larger and (b) smaller models. Each cell reports the number of games won by models of the indicated size at the corresponding position. Across ten games per family, larger models account for 18 wins and smaller models for 22.}
  \label{fig:winner_distribution2}
\end{figure*}

Fig.~\ref{fig:winner_distribution2} shows the win counts in mixed-size games (S2), grouped by model family, model size, and player position. Each game includes two larger-model agents and two smaller-model agents from the same family, with ten games evaluated per family. Panels (a) and (b) report wins by larger and smaller models, respectively, at each player position. Larger models win more games in the Qwen family, whereas smaller models win more games in the other three families. Overall, larger models account for 18 wins and smaller models for 22. No player position consistently dominates across families and model sizes.

\subsection{Game outcomes in corss-family games}
Table~\ref{tab:s3-combined} provides the detailed results of cross-family games (S3), complementing the behavioral comparison in \hyperref[sec:setup]{Section~\ref*{sec:rq2}}. The corresponding same-model results (S1) are included to examine whether the observed trading patterns persist across different opponent configurations. Consistent with the main-text observations, the ranking of trade-proposal rates remains unchanged in both size groups, whereas acceptance rates show model-specific differences.

\begin{table*}[t]
\centering
\small
\setlength{\tabcolsep}{7pt}
\begin{tabular}{lcccc}
\toprule
Model & Wins & \makecell{Wins by seat\\(P1/P2/P3/P4)}
& \makecell{Proposals per 100\\dice rolls}
& \makecell{Acceptance rate\\(\%)} \\
\midrule
\addlinespace[2pt]

Qwen3-32B & 3/20 & 0/1/1/1
& \makecell{110.0\\[-1pt]{\scriptsize (homog.\ 109.6)}}
& \makecell{54\\[-1pt]{\scriptsize (homog.\ 58)}} \\

DeepSeek-R1-32B & 2/20 & 0/1/0/1
& \makecell{92.0\\[-1pt]{\scriptsize (homog.\ 89.9)}}
& \makecell{54\\[-1pt]{\scriptsize (homog.\ 53)}} \\

Mistral-Small-24B & 1/20 & 0/0/0/1
& \makecell{86.5\\[-1pt]{\scriptsize (homog.\ 65.5)}}
& \makecell{75\\[-1pt]{\scriptsize (homog.\ 67)}} \\

Gemma4-26B & 14/20 & 3/3/3/5
& \makecell{23.3\\[-1pt]{\scriptsize (homog.\ 33.3)}}
& \makecell{56\\[-1pt]{\scriptsize (homog.\ 17)}} \\

\midrule
\addlinespace[2pt]

Qwen3-8B & 0/20 & 0/0/0/0
& \makecell{91.8\\[-1pt]{\scriptsize (homog.\ 93.6)}}
& \makecell{81\\[-1pt]{\scriptsize (homog.\ 76)}} \\

DeepSeek-R1-8B & 6/20 & 1/1/3/1
& \makecell{64.9\\[-1pt]{\scriptsize (homog.\ 73.9)}}
& \makecell{50\\[-1pt]{\scriptsize (homog.\ 33)}} \\

Mistral-Nemo-12B & 3/20 & 0/1/0/2
& \makecell{20.8\\[-1pt]{\scriptsize (homog.\ 18.1)}}
& \makecell{75\\[-1pt]{\scriptsize (homog.\ 86)}} \\

Gemma4-12B & 11/20 & 4/2/4/1
& \makecell{2.4\\[-1pt]{\scriptsize (homog.\ 1.5)}}
& \makecell{57\\[-1pt]{\scriptsize (homog.\ 55)}} \\

\bottomrule
\end{tabular}

\caption{Win counts and behavioral measures in cross-family games (S3). Each model participated in 20 games, with five games at each player position. Games lasted up to 150 steps. Trade-proposal rates and acceptance rates are pooled across the 20 games. Values in parentheses report the corresponding measures from same-model games (S1).}

\label{tab:s3-combined}
\end{table*}

\end{document}